## Construction Grammar and Language Models

Harish Tayyar Madabushi[a]*, Laurence Romain[b,c]*, Petar Milin[c], Dagmar Divjak[c]

[a] *University of Bath*
[b] *University of York*
[c] *University of Birmingham*
*\*Joint first authors*

**Abstract**:

Recent progress in deep learning and natural language processing has given rise to powerful models that are primarily trained on a cloze-like task and show some evidence of having access to substantial linguistic information, including some constructional knowledge. This groundbreaking discovery presents an exciting opportunity for a synergistic relationship between computational methods and Construction Grammar research. In this chapter, we explore three distinct approaches to the interplay between computational methods and Construction Grammar: (i) computational methods for text analysis, (ii) computational Construction Grammar, and (iii) deep learning models, with a particular focus on language models. We touch upon the first two approaches as a contextual foundation for the use of computational methods before providing an accessible, yet comprehensive overview of deep learning models, which also addresses reservations construction grammarians may have. Additionally, we delve into experiments that explore the emergence of constructionally relevant information within these models while also examining the aspects of Construction Grammar that may pose challenges for these models. This chapter aims to foster collaboration between researchers in the fields of natural language processing and Construction Grammar. By doing so, we hope to pave the way for new insights and advancements in both these fields.

Computational methods have hitherto been used by construction grammarians in one of two ways: the first is to analyse relatively large corpora to find structures which are constructional and the second, typically referred to as computational Construction Grammar, is "a branch of linguistics that aims to operationalise insights and analyses from construction grammar as computational processing models" (Van Eecke & Beuls 2018: 341). Computational methods allow for the analysis of large corpora and the automatic identification of patterns that could be considered constructions at various levels of schematicity through both syntactic and semantic analysis. This line of analysis can also be used to explain linguistic creativity, which leads to expressions such as *we will be jabbing our way out of the pandemic*, for example. Hilpert and Perek (2015) present such an application of computational methods for Construction Grammar through a robust semantic analysis of specific slots in a construction, which allows them to explain how constructions become more productive.

Recent progress in computational linguistics has led to extremely powerful deep learning models. These models not only excel on practical tasks such as machine translation and question answering, but also have access to a significant amount of linguistic information. In particular, recent work (Tayyar Madabushi et al. 2020) has shown that constructional information (i.e. potentially meaningful patterns) is available to (or 'emergent' in) deep learning models when trained on the task of predicting masked words in sentences. This finding opens up a previously unexplored synergy between computational methods and Construction Grammar research, one that is not yet widely adopted by construction grammarians due, in part, to the technical know-how and skill it demands from the user. It is the goal of this chapter to provide an overview of such deep learning models while simultaneously addressing potential reservations on the part of construction grammarians: for example, how cognitively plausible deep learning models are, how their size (i.e. the number of parameters) compares to that of human brains, and how the number of tokens they are trained on compares to typical human exposure.

In discussing the interplay between computational methods and Construction Grammar (CxG), we explicitly distinguish three approaches: (i) the use of computational methods for the analysis of text, (ii) computational CxG, which focuses on the modeling of constructional analyses, and (iii) deep learning models in computational linguistics. The opening section of this chapter offers a succinct and not comprehensive overview of the first two areas, namely computational methods and computational CxG with the aim of providing a contextual background for the recent advancements in deep learning. The second half of this chapter builds up the background necessary to gain an intuitive and linguistically grounded understanding of the workings of deep learning models, before then detailing the experiments that explore the emergence of constructionally relevant information in them. Except for the recent workshop on Construction Grammars and Natural Language Processing (Bonial & Tayyar Madabushi 2023), there is a noticeable separation between the current trajectory of research in deep learning and natural language processing and the domain of CxGs. We hope that this overview of the more traditional computational methods combined with an accessible introduction to the more recent deep learning methods will provide new and interesting avenues for collaboration between CxG and deep learning methods in computational linguistics.

## 1 The Computational Exploration of Corpora for Construction Grammar

In this section, we present an overview of research concerning the computational analysis of text within the constructional framework. We provide examples of studies where linguists incorporate computational methods for various purposes: exploring constructional meaning through specific slots, collecting constructions and organizing them into a constructicon, and training computational models to identify and reproduce constructions.

### 1.1 Vector Semantics

One branch worth mentioning is the work that has been done with vector semantics, following the well-known Firthian notion that "You shall know a word by the company it keeps" (Firth 1957: 11), which implies, among other things, that words are considered semantically similar if they tend to occur in the same contexts; i.e., there is a correlation between the co-occurrences of words and their semantic similarity.

Researchers have used various models of distributional semantics in attempts to represent word meaning. Among the first implementations of these methods are latent semantic analysis (LSA; Landauer & Dumais 1997) and hyperspace analogue to language (HAL; Lund & Burgess 1996). More recent implementations include neural network architecture such as word2vec (Mikolov et al. 2013). These methods generate representations using neural networks trained, for example, to predict a target word, given a context. The input the models are given is a representation of each word in the context, which initially consists of randomly initialized vectors and are trained to predict the target word which necessitates the models to learn an appropriate vector representation of each word. Interestingly, it was found that word2vec was mathematically equivalent to Levy and Goldberg's (2014) treatment (i.e. the factorization of co-occurrence matrices) and not significantly more powerful than earlier distributional semantic models (Levy et al. 2015).

Applications of these methods and principles to CxG became popular in the early twenty-first century. Researchers in CxG have used vector semantics models to identify constructional meaning by looking at the kind of word that occurs in a specific slot in a given construction. Levshina and Heylen (2014) explore various distributional models and clustering methods for grouping the items that occur in a specific slot of two near-synonymous Dutch causative constructions and thus identify which semantic clusters guide the choice of a particular construction. Another approach focuses on the historical development of constructions and the possible shift or extension of constructional meaning(s): Perek (2016), for example, proposed an analysis of the lexical items found in the verb slot in the [V *the hell out of*] construction to identify factors that guide syntactic productivity. Hilpert and Perek (2015) explore visualization methods for a similar purpose and offer ways of visualizing the semantic evolution of the [*many a NOUN*] construction. The main outcome from these approaches is a classification of constructions and their meanings; i.e., through advanced semantic analyses, researchers are able to identify and label subgroups of constructional meanings which can then be organized into a structured network.

## 1.2 Identifying Constructions in Building Constructicons

As it is assumed that constructions form a structured inventory rather than a list (Goldberg 1995; Hoffman & Trousdale 2013: 3), researchers have also worked towards the creation of constructicons (addressed in detail in Chapter 3), i.e. a repository of constructions that shows how they are organized in a structured network (see Chatper 9 on networks). Work on the creation of such tools is currently in progress for a variety of languages and Lyngfelt, Borin et al. (2018) offer an overview in their book *Constructicography* . To our knowledge, among the most advanced constructicons are the Russian constructicon (Janda et al. 2020) and the Swedish constructicon, with the former having been mostly constituted manually and the latter using NLP tools (cf. Lyngfelt, Borin, et al. 2018 for more details on the tools used). Indeed, for the creation of the Swedish constructicon, Lyngfelt, Bäckström et al. (2018) used computational tools to automatically extract constructions from large datasets. Their focus has mostly been on partially schematic constructions, that is, constructions that have one fixed lexical element but the surrounding slots are more schematic in that they can be filled by various lexical elements usually belonging to the same or a similar category, e.g. i_adjektivaste_laget (roughly meaning 'in the ADJ-est measure', where the adjective slot is open), but they also looked for fully schematic patterns such as the reflexive resultative (e.g. licensing expressions such as *to eat oneself full*). These schematic constructions are usually identified in corpora that are tagged for parts of speech (POS). In this way, recurring strings of POS (such as V Pn(reflexive) AP, i.e. verb – reflexive pronoun – adjective phrase) will be automatically extracted from the dataset. It then requires the intervention of a linguist to assess whether the extracted strings correspond to actual constructions. This method has the advantage of extracting different types of regularities in data and thus yielding patterns that linguists would not necessarily notice themselves, as these patterns are less obvious than "spectacular" idioms (Borin et al. 2018: 239).

More automated approaches to construction identification provide insight into frequent and thus unsurprising constructions but leave room for linguists to use a more manual approach to less frequent, unexpected constructions which can also yield relevant insights about the inner workings of schematic constructions. The automatic identification and retrieval of constructions via computational models thus holds promise, as it may help identify previously unstudied or understudied constructions, e.g., constructions with schematic structures that cannot easily be retrieved automatically from a corpus through simple queries but currently require manual annotation. In other words, well-trained computational models could also be of help in identifying constructions, as they have access to both syntactic and semantic information. An example of this is the work by Tseng et al. (2022), who begin with a popular Transformer-based language model BERT (further detailed in Section 2.2) and train it on the task of predicting masked words in variable slots of constructions. Their experiments on Taiwan Mandarin show that a model (which they call CxLM) thus trained to explicitly capture the variable slots in constructions is also capable of generating instances of constructions which are semantically and structurally valid. Similarly, work using contextual embeddings (embeddings that capture polysemy, detailed in Section 2) might hold the key to solving the issue of polysemous words, which is particularly relevant for CxG, as the various senses of polysemous words might be found in different constructions overall. For example, Romain (2022) finds that for a majority of verbs, different senses of the same verb are used in the intransitive non-causative construction and the transitive causative construction.

## 1.3 Computational Operationalizations

Fluid Construction Grammar (FCG), initially developed in 2004 and presented in Steels (2004) provides a computational operationalization of the basic tenets of CxG (discussed in detail in Chapter 21). Following one of the crucial tenets of CxG that syntax and semantics form a continuum rather than two different systems, they propose to build the syntactic and semantic aspects of linguistic structure simultaneously. Their model is bidirectional and, as the name indicates, fluid. That is, it both produces and comprehends language and it learns from every interaction. Within this branch of constructional research, van Trijp (2017) proposed a fully functional model meant to both comprehend and produce basic English, which goes beyond individual constructions. His model is quite rich lexically but limited syntactically: it knows about 35,000 lexical constructions vs. only 40 grammatical constructions. Among these 40 structures are phrasal constructions (e.g. noun phrases), argument structure constructions (intransitive, transitive, resultative, or diathetic constructions), negation patterns, and speech acts (questions vs. topicalization vs. declaratives). While the number of structures remains limited, this model is notable in that it is the first to incorporate certain crucial CxG features on a large scale; namely, it uses a frame-based approach to argument structure constructions which relies on the verb's semantic frame and its frame elements (Fillmore 1977 and also Chapter 1 in this volume). This model differs from other models in that it implements constructional language processing using a precondition-postcondition approach as a state space search process: each construction specifies a number of conditions that must be met before the construction can be applied. In addition, it comes with the FCG Editor, which is an integrated development environment for the FCG formalism, allowing for the computational verification, corpus corroboration, and integration into the applications of constructional insights (van Trijp et al. 2022).

FCG has many interesting applications. For example, Van Eecke and Beuls (2018) explored and explained how linguistic creativity occurs using this framework. As the model allows constructional productivity, e.g. the extension of the lexically fixed expression *not the sharpest tool in the box* to other semantic fields such as *not the crunchiest chip in the bag* to the schematic construction *not the X-est Y in the Z*, it also allows the free combination of multiple constructions and what they call the "appropriate violation of usual constraints" (Van Eecke & Beuls 2018). Yet another large application domain concerns experiments on emergent communication, such as, for example (Nevens et al. 2019).

## 1.4 Construction Grammar Induction Model

Dunn (2017) proposes a CxG induction algorithm which automatically identifies and extracts constructions from corpora. The model works in three different stages: the candidate generation stage, the construction identification stage, and the candidate evaluation stage (Dunn 2017: 263). In the candidate generation stage, the algorithm looks for recursive structures and non-continuous representations, e.g. *send* SOMEONE *to the cleaners*. Next, at the construction identification stage, it starts by forming templates for constructions and then looks for these constructions in the text, thus extracting potential constructions. Finally, in the third stage, the algorithm uses frequency and multi-unit association measures to select a set of likely constructions in the input corpus. The algorithm works at different levels of linguistic granularity: part of speech (using the Tree Tagger, Schmid 1994), semantic or conceptual tagging using the UCREL Semantic Analysis System (Piao et al. 2015), and identification of phrases (prepositional phrases, noun phrases) via a dependency parser (MaltParser; Nivre et al. 2007). This allows the algorithm to identify instances of the same construction whose POS structure may not be exactly the same, as in, e.g., *The coffee gave her a headache* and *The dark unfiltered coffee from South America soon gave her a splitting headache and a feeling of nausea,* which use the same ditransitive construction but whose constituents are of different sizes and structures. Through frequency and multi-unit association measures based on bi-directional ΔP, the algorithm then selects construction candidates that are sufficiently entrenched to be considered constructions. Ultimately, the model's aim is to identify actual and productive constructions from large numbers of instances of construction candidates, thus generalizing at a rather schematic level that can be extended to various substructures and instances.

Dunn presents some results from his training of the algorithm on a 1 billion word dataset from the ukWac web-crawled corpus of UK domain sites (Baroni et al. 2009). Interestingly, the algorithm summarized above manages to identify a number of structures that can be considered to share some semantic features on top of their shared syntactic features, e.g. instances of the string [Preposition] + *the* + <location>, illustrated by instances such as *on the site, in the area, into the city* or <move> + *to* + [Verb] such as *go to buy* or *come to learn*. While these structures are plausible in the sense that they have been extracted from authentic data and speakers recognize them, whether they (i) could actually be considered to be meaningful in the constructional sense and (ii) are cognitively plausible is less clear. Dunn himself acknowledges these limitations, pointing out that the algorithm uses input from a corpus, which, in this case, is a compilation of many different speakers' grammars, and his more recent work addresses some of these limitations. We will return to these considerations in Section 4, where we discuss the collaborative potential between computational approaches and CxG.

One of the main differences between FCG and Dunn's CxG induction algorithm is that Dunn's algorithm is mostly meant to show how elements that have the potential to be constructions can be automatically extracted from corpora with limited annotation. On the other hand, FCG provides a computational operationalization of the basic tenets of CxG and, thereby, provides flexible representations for constructions as well as processing and learning mechanisms for constructions. FCG, while not a model in itself, can be considered a special-purpose programming language that provides the useful abstractions and building blocks for operationalizing constructionist approaches to language.

## 2  Deep Learning and Language Models

Recent advances in computational linguistics have led to models capable of remarkable performance (Hassan et al. 2018) on specific downstream tasks, such as sentiment classification, sentence similarity, and language inference tasks. The latest deep learning based models, trained typically on thousands of examples, have outperformed non-expert crowd-sourced annotators who were asked to learn each of these tasks from a short instructions set and about 20 examples. Such models have also been shown to encode a range of linguistically relevant information, such as information about parts of speech and syntactic dependencies (Rogers et al. 2020) and aspects of constructional information, such as the ability to distinguish between sentences which are instances of the same construction and those which are not (see Sections 2.4 and 3 below). This section provides a non-technical overview of recent advances with the aim of equipping construction grammarians with the background required to appreciate the relevance of deep learning models to linguistics, and more specifically to CxGs. In particular, we hope that this will allow construction grammarians to become involved in the research pertaining to these deep learning models with an emphasis on the continued relevance of CxG.

## 2.1 From Embeddings to Pre-Trained Language Models

The need for a numeric representation of sentences which might then be fed to machine learning models as input, initially addressed through count-based embeddings, soon evolved to include neural embeddings such as word2vec. Such methods provided a way of 'encoding' text as vectors, and while this is sufficient for tasks that involve the classification of input text (e.g. sentiment classification), tasks such as machine translation and summarization require models to also output text. This is typically accomplished by the addition of a 'decoder', which generates text.

Models wherein the decoder generates text *conditioned* on the output of an encoder are called encoder-decoder models and are trained, for example, on parallel texts for translation (Sutskever 2014) and on summaries of the input text for the task of summarization. A significant shortcoming of this method is that all the information pertaining to the input text must be passed on to the decoder in the form of a single vector generated by the encoder. In complex tasks, such as translation and summarization, this was often found to result in an information bottleneck. Bahdanau et al. (2016) got around this problem through the use of 'attention weights' (model parameters) that track the importance of each input word in generating individual output words. This relatively straightforward addition of attention, it turns out, provides models with the ability to capture linguistic structure such as long-distance agreement constructions described below (Henderson 2020).

One effective way of training encoder-decoder models on large text corpora, without the need for extensive annotation, is by using language modeling. This typically consists of models being trained to predict elements of the input sequence that are 'masked' and is analogous to the cloze task. While models are often simultaneously trained on other objectives, masked language modeling is a near constant. This method of training on large text corpora is called 'pre-training', as such models can subsequently be efficiently trained ('fine-tuned') on other tasks, be it sentiment classification or summarization (Dai & Le 2015).

It has been shown that pre-training Recursive Neural Networks (RNNs; Elman 1990), specifically, Long Short-Term Memory (LSTM; Hochreiter & Schmidhuber 1997) on the language modeling objective (i.e. the prediction of masked words), enables them to capture information pertaining to linguistic phenomena such as *long-distance agreement constructions*, e.g. subject-verb agreement in English (Gulordava et al. 2018). Crucially, during training, models are not explicitly required to capture long-distance relations – they do this *incidentally* in improving their ability to predict masked words. Gulordava et al. (2018) show that this is the case even where examples are infrequent or are grammatical but meaningless (e.g. *The colourless green ideas I ate with the chair sleep furiously*) across English, Italian, Hebrew, and Russian. Such linguistic information is, in a manner of speaking, 'emergent' in pre-trained language models. This seemingly miraculous phenomenon, argued to be the result of the introduction of attention (Henderson 2020), provides an opportunity for the development of the theoretical foundations of both computational linguistics and linguistic analysis.

## 2.2 Transformers

One of the difficulties in extensive pre-training is the significant number of computational resources required for the task. The introduction of the transformer (Vaswani et al. 2017), which replaced the relatively inefficient recurrence (as in RNNs) and 'convolutions' (as in Convolutional Neural Networks or CNNs) in models with attention, significantly increased training speeds thus allowing models to be pre-trained on significantly larger corpora. This is of particular importance as it has been shown that models require a certain amount of pre-training to be able to capture interesting linguistic information and that this linguistic information is important for the models' performance on subsequent tasks on which they are fine-tuned.

The Transformer is a neural network-based sequence-to-sequence model (also known as a sequence transduction model) with an encoder-decoder architecture. The Transformer relies on modifying the attention mechanism called 'self-attention', which relates tokens in the input sequence to each other. Recall that the typical encoder-decoder model architecture had a single set of parameters that capture attention. The Transformer makes use of *several* sets of such parameters (several 'attentions' as it were, each called an attention head) so each head might capture ('attend to') different relations between words. As an example, the smaller version of BERT (detailed below) makes use of 12 attention heads in each of its 12 layers and has a total of 110 million parameters.

Several models have been developed based on the transformer architecture, the two most popular being BERT (Devlin et al. 2018) and GPT-X (Radford et al. 2018; Radford et al. 2019; Brown et al. 2020). BERT is an encoder that takes a text sequence as input and provides a numeric representation as output, while GPT-X is a decoder that takes as input a text sequence (prompt) and generates text. These models rely heavily on the pre-train-and-fine-tune paradigm and are pre-trained on very large corpora before then being fine-tuned very briefly on the specific task at hand.

Pre-training is primarily achieved through the language modeling task described above. Some models have additional training objectives, such as learning to predict if two input sentences are sequential (Devlin et al. 2018), predicting the correct order of shuffled input sentences (Lan et al. 2019), and even the correct order of words when the input words are shuffled (Clark et al. 2020). However, a systematic comparison of each of these pre-training objectives is conspicuously absent, possibly due to the extreme costs of pre-training.

Although all these models use attention, it is important to note that they are designed for performance rather than cognitive plausibility. The 16 attention heads across each of BERT Large's 24 layers for a total of 340 million parameters (Devlin et al. 2018), for example, are not intended as a parallel to the human mind. Also of possible annoyance to a linguist is the input tokenization of transformer-based pre-trained language models (PLMs). These models split less frequent words into 'subwords' not based on morphological analysis but based on frequency statistics (Sennrich et al. 2015; Kudo 2018; Church 2020). For example, *uninteresting* is split as 'un, int, ere, sting' and *hyperactive* as 'h, yper, active', which linguists would find nonsensical. Importantly, subwords and words (e.g. *active* in the above example of *hyperactive* and the word *active*) are considered to be distinct and so share no information (Gow-Smith & Tayyar Madabushi 2022).

## 2.3  Language Models and Language Learning

In exploring the significance of the *emergence* of constructional information in PLMs for the theoretical framework of CxGs, we discuss the most frequent criticisms of PLMs with respect to the concerns of Cognitive Linguistics, which includes constructional approaches: (i) the amount of pre-training data PLMs use, which is orders of magnitude greater than that which humans are exposed to, (ii) the number of parameters they have, and (iii) their lack of cognitive plausibility in their architecture and assumptions.

PLMs are trained on an inordinate amount of data: the base version of BERT, for example, is pre-trained on the BookCorpus (Zhu et al. 2015) consisting of over 11 thousand books and the entire English Wikipedia consisting of about eight billion tokens. GPT-3 is similarly trained on all of Wikipedia, the Common (web) Crawl, a second crawl of text from the web (WebText2), and two book corpora totalling about 410 billion tokens. Given that human children are exposed to between 10 and 100 million tokens (Hart & Risley 1992), models are clearly trained on a much larger amount of data. However, recent studies have shown that syntactic and semantic information is effectively captured by PLMs after training on as few as 10 to 100 million words; it is capturing world and common-sense knowledge that requires significantly more data (Zhang et al. 2020). Also, children have sensory experiences, knowledge grounding, and the benefit of attention-directed learning from their caregivers; this makes a direct comparison unfair to both parties (Linzen & Baroni 2021). PLMs have been dramatically increasing in size in terms of the parameters used, from 110M parameters in BERT base to 175 billion parameters in GPT-3. However, this is still nowhere near the human capacity at about 150 trillion synapses (Tang et al. 2001), which is the best corollary to parameters in artificial neural networks. Crucially, it is not clear that artificial neurons and neurons in the brain can be directly compared: Beniaguev et al. (2021) found that they required a 5-layer network to achieve the same level of computational capability as an L5 pyramidal neuron; the latter is the exclusive type of cells in the cerebral cortex, extending their dendrites throughout all six layers of the cortex, which positions them as a significant player in the integration of information within the outer layer of the brain responsible for higher cognitive functions. Finally, in terms of cognitive plausibility, we must consider that our understanding of the human mind is limited and that cognitive plausibility is not a prerequisite for language generation and understanding. PLMs might be moving towards an efficient language processing method that is quite distinct from humans.

## 2.4 BERTology: What Language Models 'Know'

BERTology is the recent field of research that focuses on exploring the information that transformer-based PLMs such as BERT, RoBERTa (Y. Liu et al. 2019), and GPT-X have access to. LMs, despite their focus on performance, seem to encode a surprising amount of linguistic information including constructional information. In this section, we provide an overview of the linguistic and the world and common-sense knowledge that PLMs have been shown to have access to, and then we will explore the amount of constructional information that PLMs accumulate (Section 3).

PLMs are known to have access to syntactic information, such as part of speech, constituent labeling, dependency labeling (Tenney et al. 2019), and syntactic roles (N. F. Liu et al. 2019). It has been shown that entire syntactic trees (Rosa & Mareček 2019; Kim et al. 2020;Vilares et al. 2020) as well as syntactic dependencies from the PennTreebank (Hewitt & Manning 2019) can be extracted from the representations of PLMs. Tayyar Madabushi et al. (2022) show that BERT is surprisingly good at the linguistically hard-to-define task of predicting the use of English articles. Interestingly, (Chi et al. 2020) have shown that BERT, trained on multilingual data (mBERT), has access to syntactic dependency labels, which, when represented as clusters, are consistent with the Universal Dependency taxonomy. For a detailed survey from a linguistic point of view of the syntactic capabilities of Deep Neural Networks, including models that preceded PLMs, and a discussion on its implications to linguistics, we direct the reader to the work by Linzen and Baroni (2021).

While relatively less work has gone into exploring the extent to which PLMs have access to semantic information, Tenney et al. (2019) show that PLMs seem to have information about relations, entities, and semantic roles, although multiple studies have shown that PLMs are particularly sensitive to the replacement of entities. For example, when the sentence *I thoroughly*

*enjoyed my trip to London* is modified so the location *London* is replaced by another city, BERT and similar models are prone to change the output that they assign, for example, from being classified as expressing a 'positive' sentiment about the city to the incorrect classification of being 'negative' (Balasubramanian et al. 2020; Ribeiro et al. 2020).

In addition to syntactic and semantic information, PLMs also seem to have access to world knowledge (e.g. hypernyms and hyponyms in ontologies). A complete analysis of all the linguistic and world knowledge available (and not available) to PLMs and where within the model such information might be stored is discussed in Rogers et al. (2020).

These models' access to such information should, in theory, make it easier for them to identify structures or patterns that fit the definition of a construction since they not only have access to syntactic and semantic information but also some elements of world knowledge. However, it remains unclear to what extent they are capable of combining these kinds of knowledge in a way that humans would find plausible.

## 3 The Emergence of Construction Grammar in Pre-Trained Language Models

Having provided the necessary background of pre-trained language models, we now discuss the extent to which PLMs have access to constructional information. As mentioned, pre-trained language models have been shown to have access to a significant amount of linguistic information *with no explicit training*. That is, pre-training alone results in the *emergence* of linguistic knowledge. To more comprehensively address the extent to which constructional information is emergent in PLMs, we start by exploring related cognitive linguistic phenomena, specifically polysemy and concepts (or categories).

PLMs, unlike embedding methods such as GloVe and word2vec, have been shown to learn a good estimation of the different senses of words during pre-training (Vulić et al. 2020; Garí Soler & Apidianaki 2021; Haber & Poesio 2021). In addition, significant gains have been made on the task of word sense disambiguation by the use of PLMs (Loureiro & Jorge 2019; Loureiro et al. 2022), further reinforcing this notion.

On the other hand, the differentiations among concepts in pre-trained language model representations are *not* clear-cut but rely on such information as syntax and sentiment (Yenicelik et al. 2020). For example, representations of the concept ARMS fall into different clusters based on sentiment as in the case of *handcuffed arms* (scared) and *... swooped him up into her arms* (love). This alternative approach to word senses, however, is likely to be more appealing to construction grammarians and to provide a better approximation of how humans represent meaning and concepts: "a core, tapering to a periphery" (Croft & Cruse 2004). Nair et al. (2020) similarly argue for "the potential utility of continuous-space representations of sense meanings", given their finding that there is a correlation between human judgements of the relation between meanings and distances in the BERT embeddings space.

## 3.1 Pre-Trained Language Models and Construction Grammar

The question of how much constructionally relevant information PLMs have access to, especially given that they have access to both syntactic and semantic information, has been addressed only recently, beginning with the work by Tayyar Madabushi et al. (2020). In order to assess whether BERT has access to information that could be considered constructional, Tayyar Madabushi et al. (2020) explore how the infusion of constructional information (by use of sentences which are instances of the same construction) alters BERT and how BERT performs on the task of distinguishing between sentences that instantiate the same construction(s) and those which do not. In fact, this practice of using sentences that are instances of the same construction and those which are not has continued to be a prevalent method for assessing the degree to which constructional information is encoded in language models. We provide a detailed overview of this work to illustrate methods of testing the linguistic capabilities of PLMs, as it is in the design of such experiments that the expertise of construction grammarians would most benefit computational linguists. For example, typical tasks given to speakers in behavioral studies such as sentence classification, grammaticality judgments, or cloze tasks would be great tests for models such as BERT to assess whether and to what extent it recognizes constructions. BERT, it turns out, is strikingly good at being able to distinguish between sentences that are instances of the same construction and those which are not. In fact, when trained on as few as 500 examples, BERT achieves accuracy of over 85% on this task, which is particularly surprising given that BERT was tested on a test set consisting of over 21,000 constructions (and, therefore, 21,000 independent classes). Additionally, the significantly lower performance of the GloVe BiLSTM baseline indicates that the constructions utilized may not be semantically very similar. Table 22.1 (from Tayyar Madabushi et al. 2020) shows results on how quickly BERT learns to distinguish constructions. 'Inoc' is the number of training samples used and '# of Cxns' is the number of constructions contained in that set. The rows represent the count of sentences that are instances of constructions in that set. Notice that the performance is higher on those constructions which have fewer sentences (number of sentences indicated in column one of the table) as instances, i.e., they are more restrictive.

|  | # of Cxs | Full training | Inoc 5000 | Inoc 1000 | Inoc 500 | Inoc 100 | No Training | GloVe BiLSTM Baseline |
|---|---|---|---|---|---|---|---|---|
| 2-50 | 6384 | 95.0501 | 92.8571 | 90.8130 | 88.9333 | 72.6654 | 63.9646 | 69.3163 |
| 50-100 | 2696 | 93.6573 | 92.7114 | 90.4488 | 88.9466 | 70.7901 | 63.9651 | 62.1732 |
| 100-1000 | 8974 | 94.4451 | 91.8041 | 89.4974 | 86.5612 | 60.9594 | 58.8478 | 53.9919 |
| 1000-10000 | 3266 | 89.7734 | 87.8292 | 83.2364 | 69.7336 | 57.7771 | 57.7924 | 54.1405 |
| <10000 | 21216 | 94.4075 | 90.1772 | 88.6619 | 85.7325 | 68.3352 | 61.1708 | 60.8819 |
| >10000 | 465 | 72.5498 | 72.5100 | 64.9004 | 54.9402 | 53.3865 | 52.8685 | 52.1306 |
| All | 21681 | 93.4851 | 89.3409 | 87.1339 | 85.8009 | 64.1137 | 60.1541 | 62.4242 |

Table 22.1: Evaluation of BERT Base (with no additional pre-training) on predicting if two sentences are instances of the same construction.

Also, the infusion of constructional information into BERT seems to do little to affect its ability on either downstream task, i.e., tasks that the model is subsequently fine-tuned on, or in terms of the syntactic information (e.g. PoS, Semantic Role Labeling) that it captures. This infusion of constructional information is achieved by modifying one of BERT's pre-training objectives: during pre-training, BERT is traditionally trained not only on masked language modeling (analogous to the cloze task) but also on what is called the Next Sentence Prediction task.

Tayyar Madabushi et al. (2020) pre-trained several BERT 'clones' with constructional information thus infused, using constructions that occur with different frequencies (which they call CxGBERT). For each of these frequencies, they also trained two additional BERT clones: one using the standard training data where sentences are consecutive in the training corpus (BERTClone) and the other where consecutive sentences are randomized (BERTRandom). CxGBERT and BERTClone are compared across each of the frequencies, while BERTRandom is used as a control to ensure that any differences are not an artefact of training.

The results of these experiments show that in cases where BERT is pre-trained using constructions which are less frequent, there is no significant difference between CxGBERT and BERTClone on their ability to perform on downstream tasks. However, to confirm that constructional information provides models with information comparable to that provided by sequential sentence requires a control condition. This control is provided by BERTRandom, which consistently performs worse in all but one task, indicating that the results are indeed a result of the infusion of constructional information. They also find that CxGBERT and BERTClone are very similar in the syntactic information they capture with regard to, for example, PoS, Named Entity Recognition, and Semantic Proto Labeling.

Based on these results, Tayyar Madabushi et al. (2020) conclude that BERT does in fact have access to a significant amount of information that could be assessed as constructional. Importantly, this work is not without its limitations. As mentioned by the authors, this method automatically identifies constructions and assigns sentences to them (that is, it identifies sentences that instantiate these constructions). They find, through manual analysis, that while most of these resultant groupings are indeed constructional, some of them might be too simplistic.

Li et al. (2022) similarly explore the extent to which language models have access to constructional information and more specifically, argument structure constructions using stimuli generated from templates. The authors thus adapt several psycholinguistic studies to Transformer-based language models. Following the disagreement between verb-centred approaches that claim that the verb accounts for the type of argument structure it can occur in and construction-centred approaches that assume that it is rather the construction that determines which verbs can occur in it, they test which - the verb or the construction - accounts for the lion's share of sentence meaning. Through a sentence sorting task, they find that sentences that instantiate the same argument structure construction are more closely embedded than sentences that only have the verb in common, thus supporting the constructional approach.

In an effort to simulate language acquisition (through language exposure), Li et al. (2022) use different language sample sizes and find that the more input the models get, the more likely they are to group sentences together based on their shared constructional pattern rather than their shared verb. Notably they show that RoBERTa (Y. Liu et al. 2019), for example, seems to generalize meaning without lexical overlap from different constructions. It has been argued in the acquisition literature that this also holds for learners, who seemingly progressively learn to generate from more schematic structures as they improve (Tomasello 2000; Diessel 2013, inter alia).

## 3.2 Possible Linguistic Shortcomings of Language Models

While it appears that PLMs do capture the syntactic aspects of constructions as shown by multiple studies, they seem to be failing in those aspects that require a degree of *reasoning* or *understanding*. Research into language models could potentially benefit from linguistic studies in CxG, as those generally pay attention to constructional meaning (syntax + semantics). To illustrate, we can consider work by Weissweiler et al. (2022), who study the extent to which PLMs can capture the syntactic and semantic information associated with the English comparative correlative (CC), e.g. *the better your syntax, the better your semantics* or *the more you read, the more you learn*. In evaluating the syntactic capabilities of language models, PLMs are tested for their capacity to identify sentences instantiating the CC construction through the use of 'minimal pairs', i.e., pairs of sentences that are indistinguishable except that one is an instance of the CC construction while the other is not. Experiments are conducted using both synthetic data and examples extracted from corpora. The authors generate synthetic data by modifying the second part of the CC, that is, the part that comes after '*the* X-er'. For example, sentences which are instances of the CC with the pattern '*the* ADV-er *the* NUM NOUN VERB' (*the harder the two cats fight*) are reordered as '*the* ADJ-er NUM VERB *the* NOUN" (*the harder two fight the cats*) to generate a false instance, i.e., one that is not an instance of the CC (Weissweiler et al. 2022). Sentences extracted from corpora are manually annotated using their PoS tags. These sentences are encoded using transformer based pre-trained language models (with no fine-tuning). Finally, a simple logistic regression model is trained on these encodings of sentences to distinguish between sentences which are instances of the CC and those which are not. The intuition behind this method is that the correct encoding of information pertaining to the CC within the encoding should be detectable using a simple classifier. Their results show that all the PLMs they test, namely RoBERTa, BERT, and DeBERTa (He et al. 2021) are able to distinguish between sentences which are instances of the CC and those which are not in both the synthetic and corpus data, including in instances where they involve PoS tags that had not been seen before. In this regard, their results are similar to prior work.

However, in their work exploring PLMs' ability to interpret the meaning of the CC, Weissweiler et al.'s (2022) semantic experiments using syntactic data show that the models perform poorly. Concretely, they ran usage-based experiments to evaluate if models can interpret the fact that sentences following the pattern *the* COMP .... *the* COMP imply a specific relation between the two COMPs. For example, they test to see if, given *the* ADJ1-er *you are*, *the* ADJ2-er as a pattern, the models can appropriately fill in the masked word in NAME1 *is* ADJ1-er *than* NAME2. Therefore, NAME1 *is* [MASK] *than* NAME2. While the authors made efforts to minimize biases through multiple experiments, their study is limited by the use of synthetic data for semantic evaluation. The synthetic minimal pairs employed by the authors differ quite drastically from their corpus data, raising concerns about their reliability. These concerns are further compounded by the absence of evaluation from human subjects to assess the comprehensibility of these synthetic minimal pairs. For example, the synthetic data comprises positive instances of the CC such as, *The harder and longer the three cats throw, the harder and shorter the ten dogs shake,* and negative instances such as *The higher nine strike the women without a pause the shorter ten choke the girls.*

Another potential shortcoming of the study conducted by Weissweiler et al. (2022) is that it employs relatively small models for experimentation. Recent research has demonstrated that the emergence of reasoning capabilities in PLMs occurs on a significantly larger scale, typically around 80 billion parameters (Wei et al. 2022).

## 4  Avenues for Collaboration: Computational Linguistics and Construction Grammar

Mutual enrichment between computational linguistics and CxG requires input from CxG as to the types of linguistic information that should be tested in computational models, and this is at least in part also related to issues of formalization/operationalization of the representations. A certain amount of work has already been done in the various branches of CxG to formalize constructional representations. These formalizations are potentially an excellent starting point for computational implementation as they offer detailed descriptions of constructions including syntactic structure, information about the kinds of fillers for various slots in constructions, constructional meaning, discourse-pragmatic constraints, etc.

First, of notable mention here is the vast amount of work put forward by the creators of the Russian constructicon (cf. Janda et al. 2020 for a description) who have managed to put together an inventory of 2,200 constructions.[1] In this constructicon, the constructions are organized in families, clusters, and networks, illustrated with corpus examples and even annotated for CEFR (Common European Framework of Reference for Languages) level of proficiency.

Another strand of formalization explicitly meant to be implemented computationally is Embodied Construction Grammar, which was put forward by Bergen and Chang (2005) with the aim to integrate it into a simulation-based model of language understanding. Embodied Construction Grammar follows the traditional principles of CxG but adds the extra layer of embodiment. That is, processing constructions is considered to be a full cognitive experience in that conceptual representations are grounded in the body's perceptual and motor systems. This theory therefore leads to a simplification of what needs to be represented as a construction: just enough information to activate simulations using more general sensory-motor and cognitive structures. Embodied Construction Grammar is still under development but is and has been applied to a number of issues in Cognitive Linguistics, including metaphor. Feldman (2020) provides an overview of the achievements and prospects for Embodied Construction Grammar.

## 4.1  Discovering Constructional Candidates

That being said, we have also shown that very advanced models and methods can find construction candidates in large amounts of data. For example, Li et al. (2022) have shown how models are capable of identifying and learning constructions and Dunn's induction algorithm has successfully identified some serious candidate constructions. Moreover, incorporating techniques like Abstract Meaning Representation (AMR; Knight et al. 2021) holds the potential to further enhance these capabilities. While BERTology has its own set of limiting factors that must be taken into account when interpreting these results (Rogers et al. 2020), the fact that such models are already capable of making decisions about what counts as a construction is very promising, and there is no doubt that further collaboration between construction grammarians and computational linguists can only improve these models' ability to identify, retrieve, and classify constructions. A significant step in this direction is the work conducted by Lyngfelt, Bäckström et al. (2018) in the automatic retrieval of strings/structures which they then checked manually.

---

[1] https://site.uit.no/russian-constructicon/

We have seen that advanced models such as PLMs are capable of learning and making decisions across vast amounts of data and that PLMs can handle polysemy surprisingly well. And, as shown by the work on construction and context-aware language models (CxLM), these models could achieve even more with a little bit of training in construction identification from previously annotated data. The potential for automatic annotation of vast amounts of data from smaller manually annotated samples yields very promising avenues for further research. For constructicography, this could mean the automatic identification of many instances of the same construction, which could potentially yield a large and varied array of examples used to generalize constructional meaning at various levels of granularity. Also, computational models of CxG have the potential to become very effective tools for tracking constructional productivity. Hilpert and Perek (2015) managed to identify the evolution and expansion of constructions over time through a 'simple' vector space analysis; models capable of identifying constructions in a very large dataset could thus be used to keep track of the potential productivity of constructions over periods of time. For example, they could trace and potentially predict possible extensions of constructional meanings.

## 4.2 Evidence for Usage-Based Theories

Yet another reason why collaboration between computational approaches and CxG is promising is the use of Language Models as evidence for usage-based theories of language, notably with regard to learning mechanisms. While it is tempting to dismiss advances in computational linguistics as non-human-like performance given their nature (e.g. Linzen & Baroni 2021), we would be throwing the baby with the bath water by rejecting them entirely. For a long time, linguists have st(r)ayed away from learning theory and this is mostly due to Chomsky's rebuttal of Skinner's Verbal Behavior (Skinner 1957; Chomsky 1959) and Chomsky's well-known argument about the poverty of the stimulus. Skinner argued that language was learnable by humans without the need for a language-specific apparatus and this perspective has also been adopted by cognitive linguists who consider that knowledge of language is knowledge (Goldberg 1995: 5). Now, given that a generalization engine – in this case a computational pre-trained language model - is capable of capturing a certain amount of linguistic information including constructional information when trained on lexical co-occurrences only, the question of what is learned when learning language is particularly relevant. Chi et al. (2020), for example, find that multilingual BERT can be mapped onto syntactic dependency labels, which, when represented as clusters, are consistent with Universal Dependencies.

Indeed, results from probing experiments on PLMs might support usage-based theories of language acquisition (Nivre et al. 2016). There are, however, two important aspects that must be addressed in making this claim: the first is that of sparsity of input and the second is the notion of whether or not PLMs are a tabula rasa. As mentioned in Section 2.3, despite the staggering amounts of data that PLMs are trained on, the amount of pre-training data that is required to capture syntactic and semantic phenomena is on a par with what children tend to be exposed to (Zhang et al. 2020). As such, PLMs learn the nuanced aspects of language even when exposed to input equally limited to that of children. The second argument against the idea of genetic endowment is that PLMs (Pre-Trained Language Models) are not blank slates but, rather, possess an inherent internal architecture. Changes to this architecture affect their abilities, including their capacity to capture linguistic information (Linzen & Baroni 2021). This argument suggests that any physical manifestation of a learning system would annul the tabula rasa as any system must contain some 'architecture'. Indeed, humans similarly come with cognitive abilities and predispositions that help them detect patterns in input. We contend that since PLMs' architectures (the Transformer) are not based either on language or the human mind (lack of cognitive plausibility being one of their criticisms, see Section 2.3) and their parameters are randomly initialized, they are, in fact, a blank slate.

In relation to grammar more generally, while linguistic phenomena such as parts of speech and dependencies are observed in language, it is not clear that they are anything but theoretical constructs. Clearly these phenomena do not precede language use and the use of language is not based on an a priori set of grammatical rules and patterns. It is thus possible that such linguistic phenomena are a *by-product* of language use and that they capture some — albeit complex — patterns that are the result of very different goals associated with language use, for example, with efficient communication. CxG, which emphasizes the interplay between form and meaning and the existence of a continuum between the lexical and the grammatical is well suited to capture the mechanisms underlying successful communication beyond simple syntactic rules, and is thus a great potential ally for computational linguistics as it is precisely these intricate cognitive mechanisms that language models seem to fail to mimic or grasp.

## 5 Conclusion

This chapter has focused on the nascent yet fast-evolving field involving pre-trained language models and what they capture. Recent methods in computational linguistics have led to methods for explicitly identifying constructions from corpora (e.g. the induction algorithm), but the extent to which language models capture constructional information varies. This opens up opportunities for computational linguistics and CxG to collaborate. We hope that this overview, including the background necessary to gain an intuitive and linguistically grounded understanding of the workings of deep learning models, provides a common platform for future collaboration between computational linguists and construction grammarians.